\documentclass[letterpaper, 10 pt, conference]{ieeeconf}
\IEEEoverridecommandlockouts                              
\usepackage{amsmath,amssymb,amsfonts}
\usepackage{graphicx}
\usepackage{textcomp}
\usepackage[table]{xcolor}
\usepackage{booktabs}
\usepackage{makecell}
\usepackage{multirow}
\usepackage{caption}
\usepackage{subcaption}
\usepackage{algpseudocode}
\usepackage{algorithm}
\usepackage{float}
\usepackage[hidelinks]{hyperref}
\hypersetup{
    colorlinks,
    linkcolor={red!30!black},
    citecolor={blue!30!black},
    urlcolor={blue!30!black}
}
\usepackage{diagbox}
\usepackage{lineno}
\usepackage{setspace}
\usepackage{ulem}
\usepackage{xcolor}
\usepackage{moreverb,url}
\usepackage{amssymb}

\title{\LARGE \bf
Zoom in on the Plant: Fine-grained Analysis of Leaf, Stem and Vein Instances.
}

\author{Ronja Güldenring, Rasmus Eckholdt Andersen and Lazaros Nalpantidis
\thanks{© 2023 IEEE.  Personal use of this material is permitted.  Permission from IEEE must be obtained for all other uses, in any current or future media, including reprinting/republishing this material for advertising or promotional purposes, creating new collective works, for resale or redistribution to servers or lists, or reuse of any copyrighted component of this work in other works.}%
\thanks{*This work has been supported by the European Commission and European GNSS Agency through the project ``Galileo-assisted robot to tackle the weed Rumex obtusifolius and increase the profitability and sustainability of dairy farming (GALIRUMI)", H2020-SPACE-EGNSS-2019-870258.}
\thanks{All authors are affiliated with Technical University of Denmark, Kongens Lyngby, Denmark
        {\tt\small \{ronjag, recan, lanalpa\}@dtu.dk}}%
}

\begin{document}

\maketitle
\thispagestyle{empty}
\pagestyle{empty}

\begin{abstract}
Robot perception is far from what humans are capable of. Humans do not only have a complex semantic scene understanding but also extract fine-grained intra-object properties for the salient ones. When humans look at plants, they naturally perceive the plant architecture with its individual leaves and branching system. In this work, we want to advance the granularity in plant understanding for agricultural precision robots. We develop a model to extract fine-grained phenotypic information, such as leaf-, stem-, and vein instances. The underlying dataset \textit{RumexLeaves} is made publicly available and is the first of its kind with keypoint-guided polyline annotations leading along the line from the lowest stem point along the leaf basal to the leaf apex. Furthermore, we introduce an adapted metric POKS complying with the concept of keypoint-guided polylines. In our experimental evaluation, we provide baseline results for our newly introduced dataset while showcasing the benefits of POKS over OKS.
\end{abstract}

\section{Introduction}
 In robotics, the ability to interact with the physical world is often limited by the granularity of the perception and understanding of the world. In precision agriculture, fine-grained visual analysis of individual plants can ultimately lead to the automation of labor- and physically demanding tasks, such as phenotyping or organic weed management.
%
While the automation of phenotyping in outdoor agriculture has recently shown progress in the extraction of coarse characteristics, such as the leaf shape and size~\cite{weyler2022ral, weyler2022wacv, roggiolani2023icra}, finer detail analysis, such as detecting leaf veins, has not yet been addressed. Nevertheless, the length of the veins provides relevant insights on the plant growth state and health \cite{doi:10.1021/acsomega.1c02398, https://doi.org/10.1111/pce.14225}, whereas the course of the primary vein allows one to get a more detailed plant architecture with a leaf branching pattern.
%
For weed management in crop fields, the common practice is to control weeds in early growth stages, when the joint stem is precisely identifiable. On the contrary, in grasslands, weeds only become visible in more advanced growing states when they stand out of their cluttered vegetation background. However, this comes at the cost of weeding complexity since joint stem positions are not clearly identifiable, but instead multiple precise interventions at appropriately selected points on the weed are required.
\begin{figure}[!htbp]
\centering
\begin{subfigure}{0.98\linewidth}
\centering
    \includegraphics[width=\linewidth]{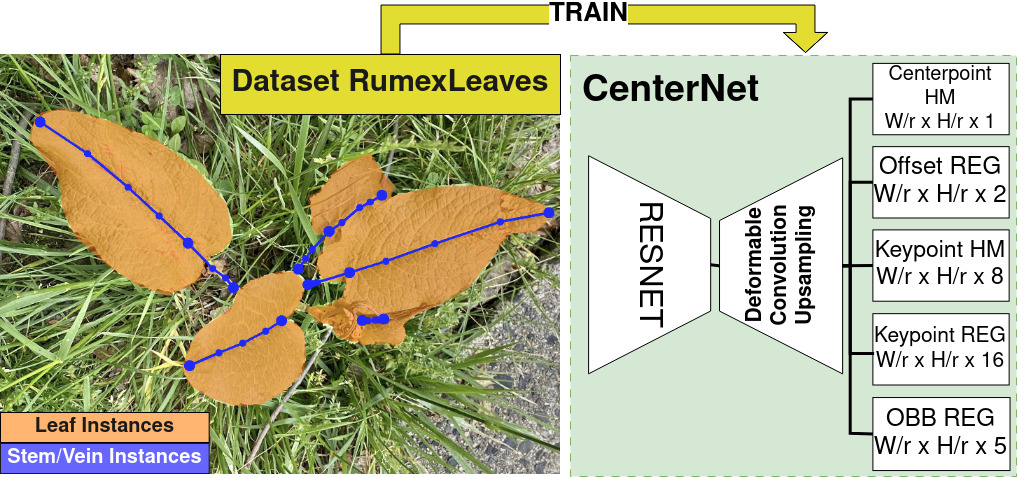}
\caption{Our public dataset \textit{RumexLeaves} includes leaf blade instance masks and keypoint-guided polylines for the stem and primary vein. We train a CenterNet model with custom heatmap (HM) and regression (REG) heads for different tasks.}
\label{fig:RumexLeaves_intro}
\end{subfigure}\\
\begin{subfigure}{.98\linewidth}
\centering
    \includegraphics[width=\linewidth]{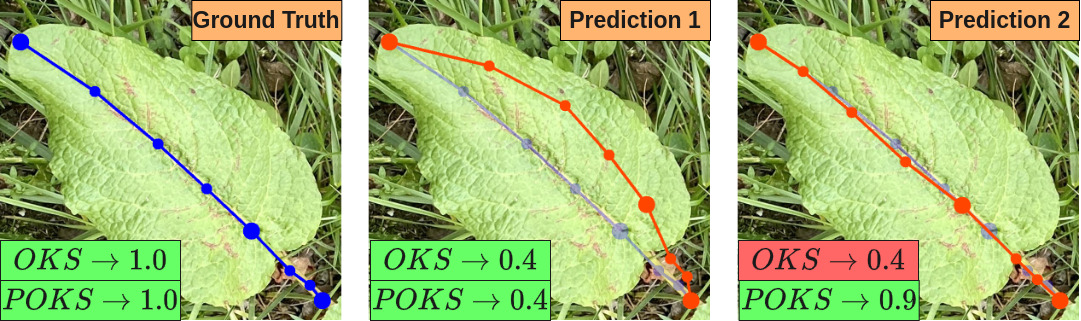}
\caption{The new metric \textit{POKS} evaluates keypoint predictions deviating from the line in accordance with the well-known OKS metric, as demonstrated in Prediction 1. However, when the keypoints deviate along the line as in Prediction 2 the metric based on \textit{POKS} remains high.}
\label{fig:POKS_intro}
\end{subfigure}
\caption{In this work we introduce (a) a new 
 publicly available dataset \textit{RumexLeaves} and (b) a new metric \textit{POKS} which complies with keypoint-guided polylines.}
\label{fig:intro_fig}
\end{figure}
\\\\
In our work, we propose an approach that extracts simultaneously leaf instances and polylines for the stem and primary vein. The polyline is guided by keypoints, whose position is distinct, such as the lowest stem point, the leaf basal, and the leaf apex. In between those distinct keypoints, pseudo keypoints are spread evenly to indicate the course of the vein, however, their position is not distinct but may deviate along the line. In order to comply with the concept of \textit{keypoint-guided polylines}, we propose an adapted metric \textit{POKS}. As illustrated in Figure~\ref{fig:POKS_intro}, \textit{POKS} evaluates predictions that deviate from the line in accordance with the well-known \textit{OKS} metric. On the contrary, when the keypoints deviate of same distance but along the line, \textit{POKS} evaluates it with higher more reasonable scores.

We showcase our work in the domain of grassland, where we extract plant traits for a highly problematic weed \textit{Rumex obtusifolius L.}~\cite{guldenring2021few, RumexWeeds}. Compared to crop fields, grasslands constitute additional challenges with more cluttered background, thin grass halms that can easily be confused with the stems as well as more occlusions from the background. Therefore, we claim that our work can be also considered for stem and vein extraction of crops of similar nature, to further push forward the automation of phenotyping.

Our contributions of this work can be summarized as followed.
\begin{itemize}
    \item We present an approach that allows for single-image fine-grained plant analysis---including stem and vein detection---in real-world settings, for the first time to the best of our knowledge. We make our code available at \url{https://github.com/DTU-PAS/RumexLeaves-CenterNet}.
    \item We introduce an adapted metric, the \textit{POKS}, that complies with keypoint-guided polylines, consisting of distinct keypoints and pseudo keypoints, which may lie at any point on neighboring line segments.
    \item We make our fine-grained dataset \textit{RumexLeaves} publicly available at \url{https://dtu-pas.github.io/RumexLeaves}, which includes 7747 manually labeled leaves with a pixel-wise annotation for the leaf blade and a polyline annotation, representing the corresponding stem and vein. To the best of our knowledge, it is the first real-world weed dataset containing stem and vein line annotations.
\end{itemize}

\section{Related Work}
\subsection{Leaf vein extraction in laboratory settings}
The extraction of leaf veins is an important property during plant phenotyping. The length and thickness of leaf veins give insights into the plant's growth state and health; the transportation capacity for water, nutrition, inorganic salts, and trace elements can be determined \cite{doi:10.1021/acsomega.1c02398, https://doi.org/10.1111/pce.14225}. Since the manual measurement of leaf veins is a time-consuming and labor-intensive task, it has been of great interest to automate the process. In recent years, automatic leaf vein segmentation has been performed with classic computer vision approaches \cite{10.1104/pp.15.00974, leaf_vein_morph, leaf_vein_sobel_hue, 10.3389/fpls.2020.00499}, traditional machine-learning approaches with hand-crafted feature extraction \cite{10.1145/3132300.3132315, leaf_vein_fft, inbook} as well as few deep learning approaches \cite{10.3389/fpls.2022.1043884}. 

In all aforementioned works, experiments are performed on sharp high-resolution (or even microscope) images in laboratory settings with controlled lighting conditions.

\subsection{Phenotyping in real-world settings}
In real-world robotics settings, joint stem detection and leaf instance detection, i.e. leaf counting, are are primarily tackled for online plant phenotyping.

Leaf counting and instance segmentation in sugar beet crop fields has been extensively studied by Weyler et al.~\cite{weyler2021ral, weyler2022ral, weyler2022wacv} and Roggiolani and Sodano et al.~\cite{roggiolani2023icra}.  Recently, they made the corresponding dataset \textit{PhenoBench}~\cite{weyler2023dataset} publicly available, which includes plant and leaf instance annotations.

When plants have a fairly small size, joint stem detection enables automatic weeding, as for example in the work of Lottes et al.~\cite{lottes18,  Lottes2020RobustFarming} and Weyler et al.~\cite{weyler2022ral}. In our previous work~\cite{agronomy13092365}, we perform joint stem detection for \textit{Rumex} plants. We found that even for the human annotator it was challenging to identify the precise joint stem position, especially in settings where plants have reached advanced growing states, their structure increases in complexity meaning the joint stems can not be clearly identified. Removing the weed gets more intractable due to the several keypoints required to fully remove them.
\\\\
In our work, we aim to perform automated leaf vein extraction in real world settings, which has not been explicitly addressed so far. While Marks et al. \cite{9811358} need to apply 3D reconstruction from multiple image views to extract advanced leaf traits, our approach only requires a single 2D image as input to solve the task of automatically extracting detailed plant traits. For \textit{Rumex} weeds in real-world grassland fields, we prove that the identification of stems, primary veins and corresponding individual leaves is feasible, advancing more fine-grained online phenotyping on robotics platforms.

\section{Methods}
\subsection{CenterNet}
The overall idea of CenterNet \cite{centernet} is to identify object instances as keypoints. An encoder-decoder network structure is used to produce a heatmap, in which each peak in the heatmap represents a detected object. Additional properties are regressed wrt. to the center point position, such as e.g. the width and height of a corresponding 2D bounding box or keypoint positions relative to the center point. Hereby, image features at the center point position are considered for the regression. CenterNet has been proven to be a solid and robust object detection approach while being easily extendable to new custom regression tasks like e.g. instance segmentation \cite{CenterMask} or multi-object tracking \cite{zhang2021fairmot}.

CenterNet performs regression on features at the spatial center point of objects. Thus, it can be infeasible to regress object properties that depend on features deviating spatially from the center point position. This especially applies to large and/or elongated objects. Deformable Convolutions (DC) \cite{deform_conv} increase the receptive field and allow the model to learn features spatially distant from the center point position. For DC, the applied image filter is not a fixed grid, but instead, the individual filter weights are applied with a learned offset. The offset depends on the input feature map and is therefore recalculated for each input tensor. The additional offset learning increases the number of model parameters and hence, the computational costs. However, common practice is to apply DC sparingly, leading to only a slight increase in model complexity.
\\\\
More details about the various architecture settings will be elucidated in analysis of Section. 
\subsection{Projected OKS}
Object Keypoint Similarity (OKS) is a popular metric to evaluate the prediction of keypoints.

\begin{equation}
\label{eq:OKS_metric}
    OKS = exp\left(-\frac{d_i^2}{2s^2\sigma_i^2}\right)
\end{equation}

As shown in Equation \ref{eq:OKS_metric}, for each keypoint,  the Euclidean distance \(d_i\) is computed between prediction and target. \(d_i\) is normalized by the scale \(s\) which corresponds to the object size as well as \(\sigma\), which represents the standard deviation between human annotations and true labels.

In this work, we aim to predict keypoint-guided polylines, which has \textit{true keypoints} with distinct positions as well as \textit{pseudo keypoints}, which are evenly spread between the \textit{true keypoints} on a certain line, however, their position is not fully distinct and can vary along the line with one degree of freedom. In our evaluation metric, we want to comply with keypoint-guided polylines and allow the \textit{pseudo keypoints} to deviate along the line to a certain extent. Therefore, we introduce a new metric, that we call \textit{Projected OKS} (POKS). 

\begin{figure}
\centering
    \includegraphics[width=0.75\linewidth]{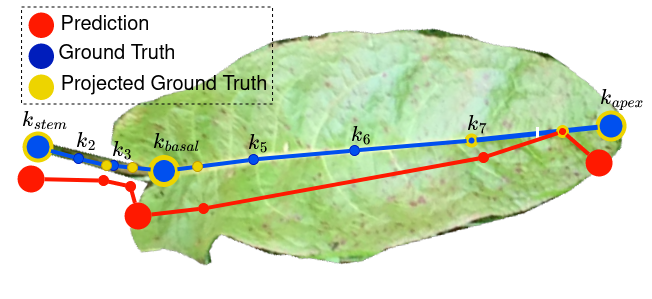}
\caption{The predicted keypoints (red) are projected to the corresponding ground truth neighbor line segments (blue). E.g. for the prediction of \(k_2\) the closest projection of the two line segments \([[k_{stem}, k_2], [k_2, k_3]]\) is retrieved. Finally, the OKS is computed between the projected points (yellow) and the predictions (red). Please note, that the red predictions are not realistic, but instead, we came up with some extreme cases to demonstrate different projection cases.}
\label{fig:keypoint_projection}
\end{figure}

While the \textit{true keypoints} are still evaluated with the standard OKS, \textit{pseudo keypoints} are projected to their two ground truth neighbor line segments. The projection serves as temporary ground truth during the evaluation. As a concrete example, please refer to Figure \ref{fig:keypoint_projection}, where the stem/vein line of a leaf is described with three \textit{true keypoints} \([k_{stem}, k_{basal}, k_{apex}]\) and five \textit{pseudo keypoints} \([k_2, k_3, k_5, k_6, k_7]\). The red keypoints represent the predictions and are projected to the corresponding ground truth segments in blue; e.g. for the prediction of  \(k_2\) the closest projection is retrieved from the two line segments \([[k_{stem}, k_2], [k_2, k_3]]\). The projection is shown in yellow and is used as ground truth to compute the \(mAP_{50:95, OKS}\). The projection is defined as follows
\begin{align}
\label{eq:projected_OKS_metric}
    p(k_i, k_x) = k_{i} + t(k_i, k_x) \cdot (k_x - k_{i}) \\
    t(k_i, k_x) = \text{clamp}\left(\frac{(\hat{y_i} - k_{i}) \cdot (k_x - k_{i})}{(k_x - k_{i}) \cdot (k_x - k_{i})}, 0, 1\right) \\
    p_{min} = \min\left(||p(k_i, k_{i-1}) - \hat{y_i}||_2, p(k_i, k_{i+1}) - \hat{y_i}||_2\right)\\
    POKS = exp\left(-\frac{p_{min}^2}{2s^2\sigma_i^2}\right)
\end{align}

where $t$ is clamped to the range $\left[0,1\right]$ to ensure the projected point falls within the line segment. Finally, the projected prediction $p_{min}$ closest to the prediction $\hat{y_i}$ is used as ground truth to compute the OKS.

\section{RumexLeaves Dataset}
\textit{RumexLeaves} contains in total 809 images with 7747 annotations of \textit{Rumex obtusifolius} leaves. For each leaf, a pixel-wise leaf blade annotation as well as a polyline annotation of the stem --- if visible --- and vein were manually created. The stem is visible for \(\approx 45 \%\) of the leaves. We differentiate between two variants of datapoints: (1) \textit{iNaturalist} datapoints that have been downloaded from the plant publisher iNaturalist and (2) \textit{RoboRumex} datapoints that have been collected with an agriculture robotics platform. Both variants originate from real-world settings and are specified further in the following subsections.
\subsection{iNaturalist}
\begin{figure*}[!htbp]
\centering
\begin{subfigure}{.74\linewidth}
\centering
    \includegraphics[width=\linewidth]{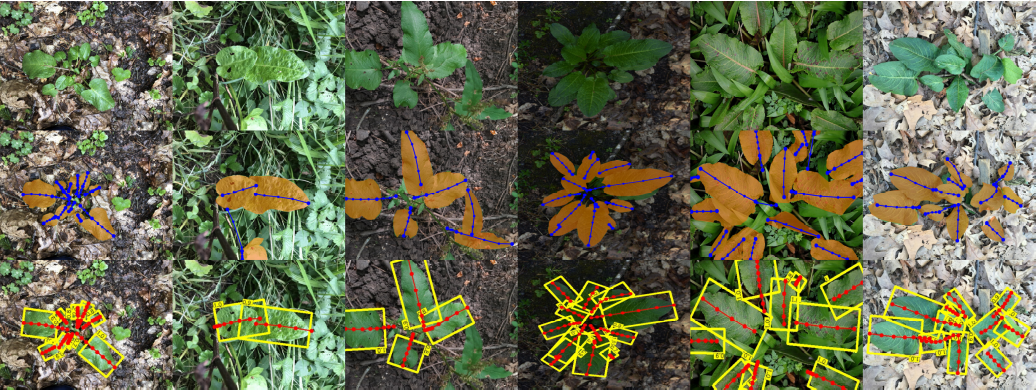}
\caption{Example iNaturalist datapoints}
\label{fig:iNaturalist_samples}
\end{subfigure}\\
\begin{subfigure}{.74\linewidth}
\centering
    \includegraphics[width=\linewidth]{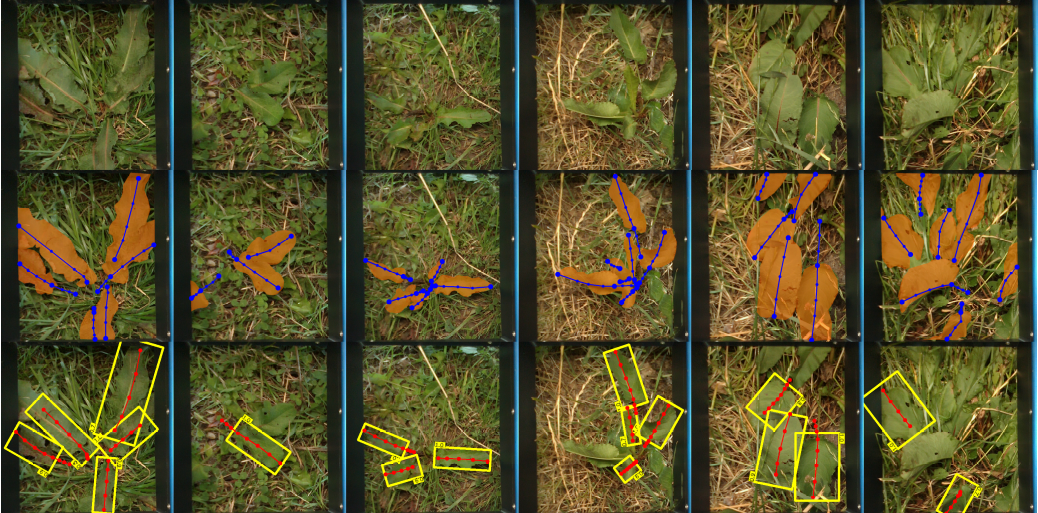}
\caption{Example RoboRumex datapoints}
\label{fig:robot_samples}
\end{subfigure}
\caption{Example RGB images (top), overlayed annotations (middle), and final predictions (bottom). Ground truth annotations are shown in orange for the leaf instance masks and blue for the stem/vein polylines. Predictions are shown in yellow for leaf instance oriented bounding bocx and in red for the stem/vein polyline.}
\label{fig:dataset_samples}
\end{figure*}

iNatualist datapoints are 690 images that have been scraped from the plant image publisher iNaturalist~\footnote{https://www.inaturalist.org/}. We additionally provide a \path{references.txt}, that includes the expected attribution as requested by \textit{CC-BY-4.0}. Since iNaturalist images are uploaded from people all over the world, the datapoints provide high variance in location, background and growing state.
Figure \ref{fig:iNaturalist_samples} provides examples for  iNaturalist datapoints.
\subsection{RoboRumex}
\begin{figure}
\centering
    \includegraphics[width=0.85\linewidth]{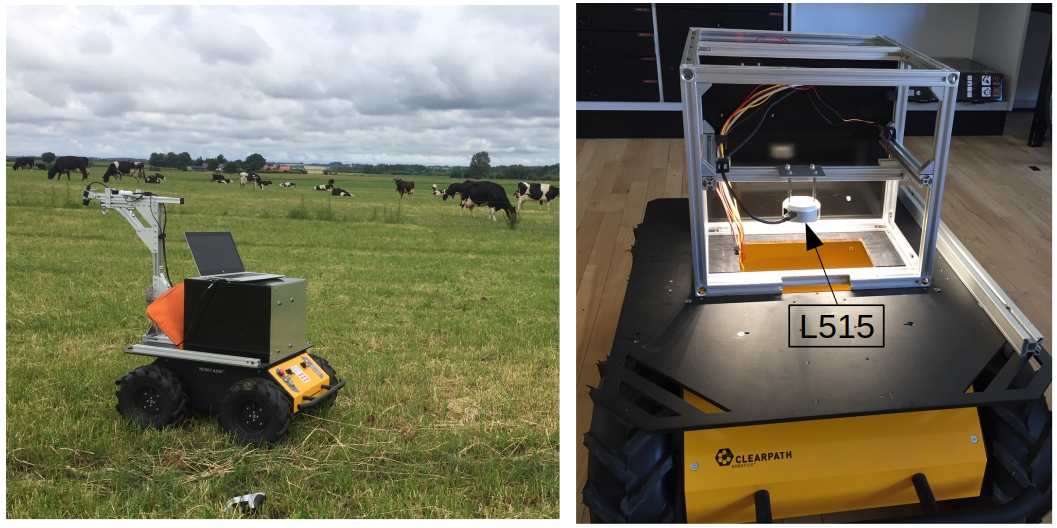}
\caption{For data collection, the Clearpath Husky Robot has been used. It is equipped with navigational sensors and two vision cameras: an RGB camera to detect the weed and an IntelRealsense L515 Lidar to further analyze the weed in a close-up view. The L515 LiDAR is mounted 0.5 m above ground pointing downward through a hole in the robot. Furthermore, the black sensor housing provides controlled lighting conditions.}
\label{fig:robot_setup}
\end{figure}

119 RoboRumex datapoints have been collected with our robot platform at three different farms in the hinterland of Copenhagen in Denmark. The vision sensor \textit{IntelRealsense L515 Lidar} is used to collect the image data, providing RGB images as well as corresponding depth maps as shown in Figure \ref{fig:robot_samples}.
Moreover, each datapoint is accompanied by navigational information such as Odometry, GNSS, and IMU of the robot.
For the data collection, we used a Clearpath Husky robot as shown in Figure \ref{fig:robot_setup}, which is specified in more detail in \cite{RumexWeeds}. 
The robot is equipped with two vision sensors: a forward-looking RGB camera in order to roughly detect the \textit{Rumex} weeds~\cite{RumexWeeds} and a close-up \textit{IntelRealsense L515 LiDAR}. Once a weed plant is detected on the RGB camera imagery, the robot drives over the plant to perform a fine-grained analysis of the plant and treat it accordingly. The \textit{L515} is mounted 0.5 m above ground pointing down through a hole in the robot chassis as shown in Figure \ref{fig:robot_setup}. Due to the \textit{L515} sharing wavelengths with natural sunlight, it is mounted within a housing which provides controlled lighting conditions.

\subsection{Image Annotations}
In order to contain a high consistency in our \textit{RumexLeaves} annotations, annotation rules have been defined beforehand.

In the following, we list the rules for the pixel-wise leaf blade annotation.\\
1. Generally, the polygon encloses the visible part of the \textit{Rumex} leaf blade. \\
2. Exception to rule 1: When leaves are occluded in such a way, that the basal and apex of the leaf is visible, occlusions must be included in the annotation.\\
3. Minor occlusions, e.g. from grass and sticks, are ignored. \\
4. If the vein of the leaf is not visible because of unfavorable angles, the leaf is not annotated. 

In the following, we list the rules for the stem and vein polyline annotation.\\
1. One polyline annotation represents the stem and the vein of one leaf with 8 keypoints: \([k_{stem}, k_2, k_3, k_{basal}, k_5, k_6, k_7, k_{apex}]\). Note, that it has 3 true keypoints \([k_{stem}, k_{basal}, k_{apex}]\) which have distinct positions. \(k_{stem}\) represents the lowest point of the stem, \(k_{basal}\) represents the lowest point of the vein at the leaf basal, and  \(k_{apex}\) represents the highest point of the vein at the leaf apex. For the remaining pseudo keypoints, the hard requirement is to place them on the stem/vein line, while a soft requirement is to roughly spread them evenly between the true keypoints in order to serve as line guidance. The ambiguous definition of those pseudo keypoints introduces strong label noise and should be taken into account during evaluation.\\
2. If the stem is not visible, the polyline only consists of 5 keypoints: \([k_{basal}, k_5, k_6, k_7, k_{apex}]\).\\
3. There is no polyline annotation if the stem only is visible.\\
4. If the basal/apex of the leaf is occluded, the lowest/highest visible point of the vein is taken.

\section{Experimental Setup}
\label{sec:exp_setup}
We develop a multi-task model, that predicts the stem/vein line as well as a rough estimate of the leaf size and orientation. However, during evaluation, we set a stronger focus on the task of stem/vein line prediction, because it is less studied by the research community and our evaluation with POKS is one of our the main contributions. Although the dataset provides segmentation masks for each leaf blade, we simplify the task and predict oriented bounding boxes (OBB), which can automatically be retrieved from the annotation: First, we use Principle Component Analysis on the vein keypoints to determine the orientation of the OBB, followed by the determination of box dimensions (width and height) on the zero oriented leaf segmentation.

\subsection{Internal Leaf Coordinate System}
\begin{figure}[!htbp]
\centering
\begin{subfigure}{.59\linewidth}
\centering
    \includegraphics[width=\linewidth]{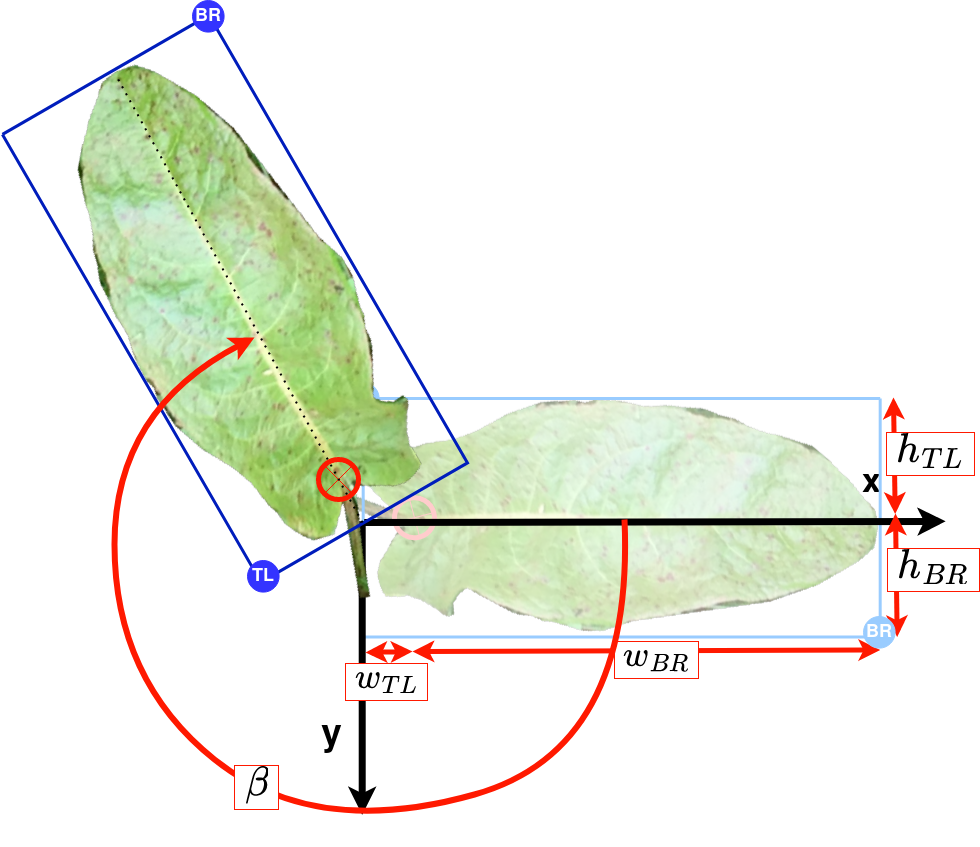}
\caption{OBB target}
\label{fig:obb_targets}
\end{subfigure}
\begin{subfigure}{.39\linewidth}
\centering
    \includegraphics[width=\linewidth]{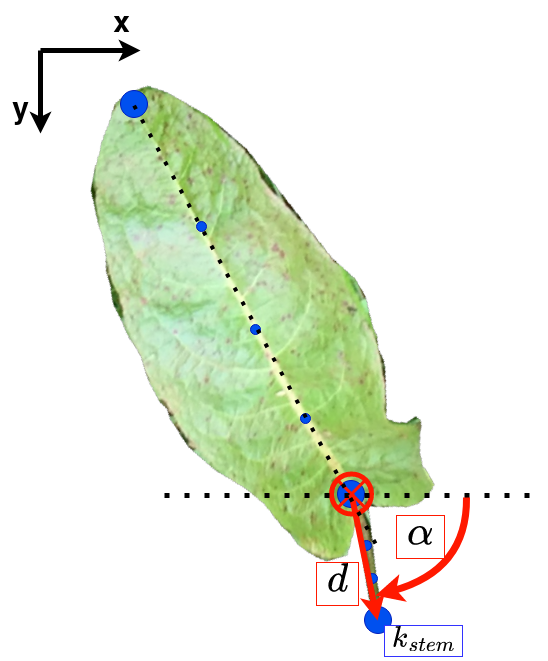}
\caption{Keypoint target}
\label{fig:kp_target}
\end{subfigure}
\caption{Keypoint and OBB targets are defined wrt. to the center point which is marked with the red circle at the leaf's basal. Left: The OBB dimensions are taken from the leaf's default position where the leaf tip is pointing along the x-axis of the image coordinate system. The angle \(\beta\) between the leaf in the default position and in the true position is the orientation of the OBB. Right: Keypoint targets are represented with the distance \(d\) and angle \(\alpha\) wrt. the center point.}
\label{fig:target_definition}
\end{figure}

During training, targets are provided in polar coordinates , because it aligns well with the regression of OBBs. For each keypoint, the distance \(d\) and angle \(\alpha\) to the center point are given as demonstrated in Figure \ref{fig:kp_target} for \(k_{stem}\). 
For the OBB prediction a straightforward approach would be to predict the width and height as well as the smallest angle \(\beta < 90°\), which points either toward the leaf's axis or perpendicular to its axis. However, we want to force the model to learn leaf hierarchies, i.e. the location of basal and apex. As shown in Figure \ref{fig:obb_targets}, we start from the zero-oriented position where the apex of the leaf is pointing in the direction of the x-axis of the image plane coordinate system. \(\beta\) is the angle between the leaf in the default position and the leaf in the true position. The width of the bounding box corresponds to the length of the leaf, while the height of the bounding box corresponds to the width of the leaf. Since the center point is not necessarily representing the center of the bounding box, but rather the object, width and height to the top-left and bottom-right corner are provided: \(w = w_{TL} + w_{BR}\), \(h = h_{TL} + h_{BR}\).

\subsection{Metrics}
We evaluate the keypoint-guided polyline with both the original OKS and the POKS introduced to show the difference. In order to get an in-depth analysis of the model performance, we report the \textit{POKS} for different parts of the polyline: \textbf{All} \([k_{stem}, k_2, k_3, k_{basal}, k_5, k_6, k_7, k_{apex}]\), \textbf{Stem} \([k_{stem}, k_2, k_3, k_{basal}]\), \textbf{Vein} \([k_{basal}, k_5, k_6, k_7, k_{apex}]\), \textbf{True} \([k_{stem}, k_{basal}, k_{apex}]\),  \textbf{Pseudo} \([k_2, k_3, k_5, k_6, k_7]\). 
\\\\
For the evaluation of the oriented Bounding Box (OBB) prediction, we consider the official implementation\footnote{https://github.com/CAPTAIN-WHU/DOTA\_devkit/blob/master/dota-v1.5\_evaluation\_task1.py} of the DOTA dataset \cite{Dota_dataset}, which is a large-scale benchmark dataset for oriented object detection in aerial images. The implementation is based on the original \(mAP_{50}\) VOC implementation for straight bounding boxes.

\subsection{Model Design}
In Figure~\ref{fig:RumexLeaves_intro}, we give an overview of the model architecture with all potential heatmap (HM) and regression (REG) heads. The architecture has a shared encoder-decoder block, consisting of the ResNet architecture as backbone, followed by three upsampling blocks with 256, 128, and 64 output channels respectively. On top of that, multiple heads are considered during our experimental evaluation as described in the following.
\begin{itemize}
    \item Centerpoint heatmap \([W/r \times H/r \times 1]\): Each peak in the heatmap represents a detected leaf. The output is downscaled by ratio \(r\).
    \item Offset regression \([W/r \times H/r \times 2]\): The identified center point position of each leaf is corrected by the offset in order to comply to the down-scaled output map.
    \item Keypoint regression \([W/r \times H/r \times 16]\): 8 keypoints \((d, \alpha\)) are regressed wrt. the corresponding centerpoint.
    \item Keypoint heatmap \([W/r \times H/r \times 7/8]\): The keypoint prediction is refined by assigning each regressed keypoint location to the closest keypoint in the heatmap \cite{centernet}. We consider two different type of heatmaps: P-heatmap and S-heatmap. P-heatmap has 8 channels and is a Gaussian rendered heatmap based on the keypoint position. S-heatmap has 7 channels is a Gaussian rendered heatmap based on the polyline segments, i.e. \([k_{stem}, k_2], [k_2, k_3], ... , [k_7, k_{apex}]\). 
    \item OBB regression \([W/r \times H/r \times 3/5]\): Either 3 \((w, h, \beta)\) or 5 \((w_{TL}, w_{BR}, h_{TL}, h_{BR}, \beta)\) are regressed wrt. the corresponding centerpoint.
\end{itemize}
The final loss consists of a weighted sum of L1 loss for regression heads and focal loss for dense object detection \cite{8237586} for heatmap heads as follows
\begin{equation*}
    \mathcal{L}_{total} = w_{cp} \mathcal{L}_{f} + w_{off} \mathcal{L}_{1} + w_{kp} \mathcal{L}_{1} + w_{kphm} \mathcal{L}_{f} + w_{obb} \mathcal{L}_{1}
\label{eq:loss}
\end{equation*}
, where \(w_{cp} = 1.0\), \(w_{off} = 1.0\), \(w_{kp} = 20.0\), \(w_{kphm} = 1.0\), \(w_{obb} = 20.0\).

Note, that we do not apply any post-processing of predictions, such as restricting the predicted keypoints to the boundaries of the predicted OBB.

\subsection{Fixed Hyperparameters}
For all experiments, we fix a number of training settings. During our ablation studies, we only consider the iNaturalist images from \textit{RumexLeaves}, which are split randomly into train, val, and test by 0.7/0.15/0.15. The input size is fixed to \((512, 512)\) to ensure that the stems of the leaves are represented with a large enough amount of pixels to be distinguishable and identifiable. All models are trained with a batch size of 32 for 96000 iterations, which corresponds to 6001 epochs for the iNaturalist train split. The network is updated with the Adam optimizer, with multistep learning scheduling applied. The initial learning rate is at 0.0005 and is updated at epoch \([2500, 4000, 5000]\) with a constant \(\gamma = 0.5\). During training, standard augmentation is applied to the image data: random flip, zoom out, color jitter, rotate by 90°, random brightness, random contrast. Furthermore, the images are normalized. The model backbone is initialized with pre-trained ImageNet weights. 

\section{Experimental Results}
\label{sec:exp_results}
In section \ref{sec:ablation_studies}, we perform ablation studies for each task -- keypoint and OBB prediction -- individually. In section \ref{sec:multi_task_performance}, we fix the settings to the best results of the ablation studies and train a model in a multi-task setting. The best-performing model is applied to the test data split in section \ref{sec:final_results}, accompanied by some qualitative prediction examples.
\subsection{Ablation Studies}
\label{sec:ablation_studies}
\subsubsection{Target Design}
\label{sec:target_design}
In this ablation study, the target design for the keypoint prediction only is examined in Table \ref{tab:abl_target}. 

In the first two experiments, we elaborate on the angle representation of the keypoint prediction. In experiment 0, the angle is predicted as normalized angle \(\alpha\) in the range of \([0, 1]\). In experiment 1, we use the representation \(cos(\alpha), sin(\alpha)\) which ensures continuity throughout the whole angle range. The value pair can be decoded back to \(\alpha = arctan(sin(\alpha), cos(\alpha))\). The results in Table \ref{tab:abl_target} show, that experiment 1 with continuous angle representation improves over experiment 0. Therefore, we fix this representation for the following experiments.

In the original CenterNet \cite{centernet} implementation, the center point is represented by the center of the object's bounding box. In order to increase the focus on the stem and vein line, the true keypoints on the polyline are considered as center points to represent a leaf. Experiments 2 -- 5 in Table \ref{tab:abl_target} show the results for the different center points and using \(k_{basal}\) as center point clearly outperforms its alternatives. Therefore, we use \(k_{basal}\) as center point for all following experiments.

Experiments 6 -- 7 in Table \ref{tab:abl_target} show that introducing heatmap refinement -- via P-heatmap or S-heatmap -- leads to some improvements. However, the difference between P-heatmap and S-heatmap is not crucial.

\begin{table*}
    \caption{Ablation study for different target designs.}
    \centering
    \begin{tabular}{c|c|c|c|c|c|c|c|c}
    \toprule 
    & \multicolumn{2}{c|}{Target}& \multicolumn{5}{c|}{\(mAP_{50:95, POKS}\) } & \(mAP_{50:95, OKS}\) \\
    ID & Which target? & Specification & All & Stem & Vein & True & Pseudo & \\
    0 & \multirow{2}{*}{angle} & angle/360 & 0.354 & \underline{0.280} & 0.448 & 0.167 & 0.481 & 0.186 \\
    1 &  & cos(angle), sin(angle) & \underline{0.378} & 0.275 & \underline{0.491} & \underline{0.175} & \underline{0.514} & \underline{0.196} \\
    \midrule
    2 & \multirow{4}{*}{center point} & center of OBB & 0.313 & 0.165 & 0.474 & 0.132 & 0.442 & 0.146 \\
    3 &  & \(kp_{stem}\) & 0.287 & 0.279 & 0.299 & 0.133 & 0.405 & 0.150 \\
    4 &  & \(kp_{basal}\) & \underline{0.378} & \underline{0.275} & \underline{0.491} & \underline{0.175} & \underline{0.514} & \underline{0.196} \\
    5 &  & \(kp_{apex}\) & 0.320 & 0.130 & 0.505 & 0.145 & 0.436 & 0.138 \\
    \midrule
    6 & \multirow{4}{*}{keypoints} & regressed only & 0.378 & 0.275 & 0.491 & 0.175 & 0.514 & 0.196 \\
    7 & & regressed + P-heatmap & 0.382 & 0.291 & \textbf{\underline{0.493}} & 0.182  & \textbf{\underline{0.521}} & 0.203 \\
    8 & & regressed + S-heatmap & \textbf{\underline{0.385}} & \textbf{\underline{0.302}} & 0.483 & \textbf{\underline{0.185}} & 0.519 & \textbf{\underline{0.208}}\\
    \bottomrule
    \end{tabular}
\center
\label{tab:abl_target}
\end{table*}

As expected, the \(mAP_{50:95, POKS}\) performs better than the original \(mAP_{50:95, OKS}\) due to the looser projection criterion. The performance of the stem prediction is clearly lower than the one of the veins. Stems are often camouflaged in the background, which has similar characteristics such as grass stalks or small sticks. Moreover, stems contain only a small amount of pixels, which decreases the number of available features. One could argue, that veins are as thin as stems and are therefore represented by the same number of features. However, veins are embedded in the leaf blade which indirectly inflates the number of features, leading to a better prediction performance.

\subsubsection{Model Complexity}
\label{sec:model_complexity}
We take the target setup from experiment 4 in Table \ref{tab:abl_target} and vary the models complexity in Table \ref{tab:abl_complexity_kp}. In experiment 0, a model with a ResNet18 backbone and two Convolutional Layers for the keypoint head is used. When looking at the model predictions qualitatively, it is striking that keypoint predictions close to the center point are more accurate than those that deviate from the center point. The reason for a decreasing performance with increasing distance from the center point is a potentially too small receptive field that is not capable of taking into account features further away from the center point. In experiment 1, we increase the receptive field by increasing the number of layers. This action has minimal impact on the performance of the network. In experiment 2, the Convolutional Layers in the up-sampling blocks of the ResNet are replaced with Deformable Convolutions, leading to a relevant performance boost with 2.8\% improvement for the \(mAP_{50:95, POKS}\) and 6\% improvement for the original \(mAP_{50:95, OKS}\). Finally, we replace the Convolutional Layers in the keypoint head with Deformable Convolution, leading to further improvement. The best model is a Resnet backbone with Deformable Convolutions in the upsampling blocks as well as two Deformable Convolutions in the keypoint head as listed in experiment 3. In experiments 5 and 6, the backbone complexity is further increased by applying deeper models, such as ResNet34 and ResNet50, resulting in further improvement. The best performing model is the ResNet50 variant with a \(mAP_{50:95, POKS}\) of 0.460 and a \(mAP_{50:95, OKS}\) of 0.267.
\begin{table*}
    \caption{Ablation study of model complexities for keypoint detection.}
    \centering
    \begin{tabular}{c|c|c|c|c|c|c|c|c}
    \toprule 
    ID & \multicolumn{2}{c|}{Model Complexity} & \multicolumn{5}{c|}{\(mAP_{50:95, POKS}\) } & \(mAP_{50:95, OKS}\) \\
    & Backbone & KP Head & All & Stem & Vein & True & Pseudo & \\
    \midrule
    0 & ResNet18 & 2 Conv & 0.350 & 0.282 & 0.432 & 0.136 & 0.500 & 0.170 \\
    1 & ResNet18 & 3 Conv & 0.352 & 0.292 & 0.429 & 0.138 & 0.501 & 0.179 \\
    2 & ResNet18 + DC & 2 Conv & 0.378 & 0.275 & 0.491 & 0.175 & 0.514 & 0.196 \\
    3 & ResNet18 + DC & 2 DC & \underline{0.405} & \underline{0.301} & \underline{0.522} & \underline{0.219} & \underline{0.528} & \underline{0.230} \\
    4 & ResNet18 + DC & 3 DC & 0.397 & 0.295 & 0.514 & 0.217 & 0.517 & 0.224\\
    \midrule
    5 & ResNet34 + DC & 2 DC & 0.433 & 0.325 & 0.549 & 0.237 & 0.556 & 0.242\\
    6 & ResNet50 + DC & 2 DC  & \textbf{\underline{0.460}} & \textbf{\underline{0.366}} & \textbf{\underline{0.563}} & \textbf{\underline{0.259}} & \textbf{\underline{0.586}} & \textbf{\underline{0.267}}\\
    \bottomrule
    \end{tabular}
\label{tab:abl_complexity_kp}
\end{table*}

We apply the same ablation study of model complexities for OBB detection only by introducing Deformable Convolutions in the upsampling blocks of the ResNet as well as in the OBB head. The results are shown in Table \ref{tab:abl_complexity_obb} and the effect of Deformable Convolutions is not as strong, yet visible. The best-performing variant -- ResNet18 + DC and 3 Deformable Convolutions in the OBB head  -- is further used with deeper ResNet variants.

\begin{table}
    \caption{Ablation study of model complexities for OBB detection.}
    \centering
    \begin{tabular}{c|c|c|c}
    \toprule 
    ID & Backbone & OBB head & \(mAP_{50, OBB}\)\\
    \midrule
    0 & ResNet18 & 2 Conv & 0.705\\
    1 & ResNet18 & 3 Conv & 0.706\\
    2 & ResNet18 + DC & 2 Conv & 0.709\\
    3 & ResNet18 + DC & 2 DC & 0.710\\
    4 & ResNet18 + DC & 3 DC & \underline{0.712}\\
    \midrule
    5 & ResNet34 + DC & 3 DC & 0.714\\
    6 & ResNet50 + DC & 3 DC & \textbf{0.776}\\
    \bottomrule
    \end{tabular}
\label{tab:abl_complexity_obb}
\end{table}

\subsection{Multi-Task Performance}
\label{sec:multi_task_performance}
We apply the best performing model 6 from Table \ref{tab:abl_complexity_kp} to a multi-task setting where both tasks - keypoint and OBB detection -- are solved simultaneously. Results are listed in Table \ref{tab:multi_task_perf}. We used the loss weights \(w_{obb} = 10.0\) and \(w_{kp} = 10.0\). Though the individual models marginally outperform the multi-task model, the performance drop is not large enough to justify the increased computational complexity.
\begin{table*}
    \caption{Multi-task performance wrt. to their single task performance (\(\downarrow\)).}
    \centering
    \begin{tabular}{c|c|c|c|c|c|c}
    \toprule 
    \multicolumn{5}{c|}{\(mAP_{50:95, POKS}\) } & \(mAP_{50:95, OKS}\) & \(mAP_{50, OBB}\) \\
     All & Stem & Vein & True & Pseudo & & \\
    \midrule
    \makecell{0.446 \\ \((\downarrow 0.014)\)} & \makecell{0.347 \\ \((\downarrow 0.019)\)} & \makecell{0.559 \\ \((\downarrow 0.004)\)} & \makecell{0.246 \\ \((\downarrow 0.013)\)} & \makecell{0.574 \\ \((\downarrow 0.012)\)} & \makecell{0.256 \\ \((\downarrow 0.011)\)} & \makecell{0.709 \\ \((\downarrow 0.067)\)}\\
        \bottomrule
    \end{tabular}
\label{tab:multi_task_perf}
\end{table*}

\subsection{Final Results}
\label{sec:final_results}
The multi-task model from the previous section serves as our final model. We apply the final model to the iNaturalist as well as the RoboRumex datapoints and list the results in Table
\ref{tab:test_results}. Note, that the model is trained on 70\% of the iNaturalist datapoints only. The model performs on par on the iNaturalist test split, but much worse on the RoboRumex test split. We evaluate on the RoboRumex test split only to allow comparisons with future developments, that potentially include RoboRumex data during training. Although the Rumex leaves in the iNaturalist datapoints are diverse with a range of different backgrounds and different leaf shapes as well as appearances, the model is not performing sufficiently on a new data distribution such as the one collected with the robot. Reasons for the lower performance can be the image quality/sharpness as well as the predominantly occurrence of small leaves.
\begin{table*}
    \caption{Final performance on test split of iNaturalist and RoboRumex datapoints.}
    \centering
    \begin{tabular}{c|c|c|c|c|c|c|c|c}
    \toprule 
    ID & Test Split & \multicolumn{5}{c|}{\(mAP_{50:95, POKS}\) } & \(mAP_{50:95, OKS}\) & \(mAP_{50, OBB}\) \\
     &  &  All & Stem & Vein & True & Pseudo & & \\
    \midrule
    0 & iNaturalist & 0.450 & 0.313 & 0.600 & 0.246 & 0.579 & 0.265 & 0.716 \\
    1 & RoboRumex &  0.244 & 0.171 & 0.390 & 0.080 & 0.406 & 0.109 & 0.669 \\
    \bottomrule
    \end{tabular}
\label{tab:test_results}
\end{table*}
In Figure \ref{fig:iNaturalist_samples} and \ref{fig:robot_samples}, we show the prediction vs. ground truth for sample images for the iNaturalist and RoboRumex test split datapoints respectively.

\section{Conclusion}
We present a fine-grained plant analysis in a real-world grassland setting. We train a custom CenterNet model to extract each leaf instance with an oriented bounding box and the corresponding stem and vein polyline. The performance is evaluated with standard metrics such as OKS and mAP as well as an adapted keypoint evaluation \textit{POKS}, which complies with keypoint-guided polylines, consisting of distinct as well as pseudo keypoints. Additionally, we publish the corresponding dataset \textit{RumexLeaves} with 7747 manually annotated leaves at \url{https://dtu-pas.github.io/RumexLeaves} to encourage other researchers to improve on our results.  

\bibliographystyle{IEEEtran}
\bibliography{IEEEabrv, references}

\begin{thebibliography}{10}
\providecommand{\url}[1]{#1}
\csname url@rmstyle\endcsname
\providecommand{\newblock}{\relax}
\providecommand{\bibinfo}[2]{#2}
\providecommand\BIBentrySTDinterwordspacing{\spaceskip=0pt\relax}
\providecommand\BIBentryALTinterwordstretchfactor{4}
\providecommand\BIBentryALTinterwordspacing{\spaceskip=\fontdimen2\font plus
\BIBentryALTinterwordstretchfactor\fontdimen3\font minus \fontdimen4\font\relax}
\providecommand\BIBforeignlanguage[2]{{%
\expandafter\ifx\csname l@#1\endcsname\relax
\typeout{** WARNING: IEEEtran.bst: No hyphenation pattern has been}%
\typeout{** loaded for the language `#1'. Using the pattern for}%
\typeout{** the default language instead.}%
\else
\language=\csname l@#1\endcsname
\fi
#2}}

\bibitem{weyler2022ral}
J.~Weyler, J.~Quakernack, P.~Lottes, J.~Behley, and C.~Stachniss, ``{Joint Plant and Leaf Instance Segmentation on Field-Scale UAV Imagery},'' \emph{IEEE Robotics and Automation Letters}, vol.~7, no.~2, pp. 3787--3794, 2022.

\bibitem{weyler2022wacv}
J.~Weyler, F.~Magistri, P.~Seitz, J.~Behley, and C.~Stachniss, ``{In-Field Phenotyping Based on Crop Leaf and Plant Instance Segmentation},'' in \emph{IEEE/CVF Winter Conference on Applications of Computer Vision}, 2022.

\bibitem{roggiolani2023icra}
G.~Roggiolani, M.~Sodano, T.~Guadagnino, F.~Magistri, J.~Behley, and C.~Stachniss, ``Hierarchical approach for joint semantic, plant instance, and leaf instance segmentation in the agricultural domain,'' in \emph{Proc. of the IEEE International Conference on Robotics and Automation (ICRA)}, 2023.

\bibitem{doi:10.1021/acsomega.1c02398}
\BIBentryALTinterwordspacing
T.~J. Konch, T.~Dutta, M.~Buragohain, and K.~Raidongia, ``{Remarkable Rate of Water Evaporation through Naked Veins of Natural Tree Leaves},'' \emph{ACS Omega}, vol.~6, no.~31, pp. 20\,379--20\,387, 2021. [Online]. Available: \url{https://doi.org/10.1021/acsomega.1c02398}
\BIBentrySTDinterwordspacing

\bibitem{https://doi.org/10.1111/pce.14225}
\BIBentryALTinterwordspacing
L.~Pan, B.~George-Jaeggli, A.~Borrell, D.~Jordan, F.~Koller, Y.~Al-Salman, O.~Ghannoum, and F.~J. Cano, ``Coordination of stomata and vein patterns with leaf width underpins water-use efficiency in a c4 crop,'' \emph{Plant, Cell \& Environment}, vol.~45, no.~6, pp. 1612--1630, 2022. [Online]. Available: \url{https://onlinelibrary.wiley.com/doi/abs/10.1111/pce.14225}
\BIBentrySTDinterwordspacing

\bibitem{guldenring2021few}
R.~G{\"u}ldenring, E.~Boukas, O.~Ravn, and L.~Nalpantidis, ``Few-leaf learning: Weed segmentation in grasslands,'' in \emph{2021 IEEE/RSJ International Conference on Intelligent Robots and Systems (IROS)}.\hskip 1em plus 0.5em minus 0.4em\relax IEEE, 2021, pp. 3248--3254.

\bibitem{RumexWeeds}
\BIBentryALTinterwordspacing
R.~Güldenring, F.~K. van Evert, and L.~Nalpantidis, ``{RumexWeeds}: A grassland dataset for agricultural robotics,'' \emph{Journal of Field Robotics}. [Online]. Available: \url{https://onlinelibrary.wiley.com/doi/abs/10.1002/rob.22196}
\BIBentrySTDinterwordspacing

\bibitem{10.1104/pp.15.00974}
\BIBentryALTinterwordspacing
J.~Bühler, L.~Rishmawi, D.~Pflugfelder, G.~Huber, H.~Scharr, M.~Hülskamp, M.~Koornneef, U.~Schurr, and S.~Jahnke, ``{phenoVein—A Tool for Leaf Vein Segmentation and Analysis  },'' \emph{Plant Physiology}, vol. 169, no.~4, pp. 2359--2370, 10 2015. [Online]. Available: \url{https://doi.org/10.1104/pp.15.00974}
\BIBentrySTDinterwordspacing

\bibitem{leaf_vein_morph}
X.~Zheng and X.~Wang, ``Leaf vein extraction based on gray-scale morphology,'' \emph{International Journal of Image, Graphics and Signal Processing}, vol.~2, 12 2010.

\bibitem{leaf_vein_sobel_hue}
C.~Li, C.~Sun, J.~Wang, and F.~Li, ``Extraction of leaf vein based on improved sobel algorithm and hue information,'' \emph{Nongye Gongcheng Xuebao/Transactions of the Chinese Society of Agricultural Engineering}, vol.~27, pp. 196--199, 07 2011.

\bibitem{10.3389/fpls.2020.00499}
\BIBentryALTinterwordspacing
J.~Zhu, J.~Yao, Q.~Yu, W.~He, C.~Xu, G.~Qin, Q.~Zhu, D.~Fan, and H.~Zhu, ``A fast and automatic method for leaf vein network extraction and vein density measurement based on object-oriented classification,'' \emph{Frontiers in Plant Science}, vol.~11, 2020. [Online]. Available: \url{https://www.frontiersin.org/articles/10.3389/fpls.2020.00499}
\BIBentrySTDinterwordspacing

\bibitem{10.1145/3132300.3132315}
\BIBentryALTinterwordspacing
J.~D.~S. Selda, R.~M.~R. Ellera, L.~C. Cajayon, and N.~B. Linsangan, ``Plant identification by image processing of leaf veins,'' in \emph{Proceedings of the International Conference on Imaging, Signal Processing and Communication}, ser. ICISPC 2017.\hskip 1em plus 0.5em minus 0.4em\relax New York, NY, USA: Association for Computing Machinery, 2017, p. 40–44. [Online]. Available: \url{https://doi-org.proxy.findit.cvt.dk/10.1145/3132300.3132315}
\BIBentrySTDinterwordspacing

\bibitem{leaf_vein_fft}
K.~Lee and K.-S. Hong, ``An implementation of leaf recognition system using leaf vein and shape,'' \emph{International Journal of Bio-Science and Bio-Technology}, vol.~5, pp. 57--65, 01 2013.

\bibitem{inbook}
G.~Samanta, A.~Chakrabarti, and B.~Bhattacharya, \emph{Extraction of Leaf-Vein Parameters and Classification of Plants Using Machine Learning}, 01 2021, pp. 579--586.

\bibitem{10.3389/fpls.2022.1043884}
\BIBentryALTinterwordspacing
X.~Liu, B.~Xu, W.~Gu, Y.~Yin, and H.~Wang, ``Plant leaf veins coupling feature representation and measurement method based on deeplabv3+,'' \emph{Frontiers in Plant Science}, vol.~13, 2022. [Online]. Available: \url{https://www.frontiersin.org/articles/10.3389/fpls.2022.1043884}
\BIBentrySTDinterwordspacing

\bibitem{weyler2021ral}
J.~Weyler, A.~Milioto, T.~Falck, J.~Behley, and C.~Stachniss, ``{Joint Plant Instance Detection and Leaf Count Estimation for In-Field Plant Phenotyping},'' \emph{IEEE Robotics and Automation Letters}, vol.~6, pp. 3599--3606, 2021.

\bibitem{weyler2023dataset}
J.~Weyler, F.~Magistri, E.~Marks, Y.~L. Chong, M.~Sodano, G.~Roggiolani, N.~Chebrolu, C.~Stachniss, and J.~Behley, ``{PhenoBench --- A Large Dataset and Benchmarks for Semantic Image Interpretation in the Agricultural Domain},'' \emph{arXiv preprint}, 2023.

\bibitem{lottes18}
P.~Lottes, J.~Behley, N.~Chebrolu, A.~Milioto, and C.~Stachniss, ``Joint stem detection and crop-weed classification for plant-specific treatment in precision farming,'' in \emph{IEEE/RSJ International Conference on Intelligent Robots and Systems (iROS)}, 06 2018.

\bibitem{Lottes2020RobustFarming}
------, ``Robust joint stem detection and crop-weed classification using image sequences for plant-specific treatment in precision farming,'' \emph{Journal of Field Robotics}, vol.~37, no.~1, pp. 20--34, 2020.

\bibitem{agronomy13092365}
\BIBentryALTinterwordspacing
J.~Li, R.~Güldenring, and L.~Nalpantidis, ``Real-time joint-stem prediction for agricultural robots in grasslands using multi-task learning,'' \emph{Agronomy}, vol.~13, no.~9, 2023. [Online]. Available: \url{https://www.mdpi.com/2073-4395/13/9/2365}
\BIBentrySTDinterwordspacing

\bibitem{9811358}
E.~Marks, F.~Magistri, and C.~Stachniss, ``Precise 3d reconstruction of plants from uav imagery combining bundle adjustment and template matching,'' in \emph{2022 International Conference on Robotics and Automation (ICRA)}, 2022, pp. 2259--2265.

\bibitem{centernet}
\BIBentryALTinterwordspacing
X.~Zhou, D.~Wang, and P.~Kr{\"{a}}henb{\"{u}}hl, ``Objects as points,'' \emph{CoRR}, vol. abs/1904.07850, 2019. [Online]. Available: \url{http://arxiv.org/abs/1904.07850}
\BIBentrySTDinterwordspacing

\bibitem{CenterMask}
Y.~Wang, Z.~Xu, H.~Shen, B.~Cheng, and L.~Yang, ``Centermask: Single shot instance segmentation with point representation,'' in \emph{2020 {IEEE/CVF} Conference on Computer Vision and Pattern Recognition, {CVPR} 2020, Seattle, WA, USA, June 13-19, 2020}.\hskip 1em plus 0.5em minus 0.4em\relax Computer Vision Foundation / {IEEE}, 2020, pp. 9310--9318.

\bibitem{zhang2021fairmot}
Y.~Zhang, C.~Wang, X.~Wang, W.~Zeng, and W.~Liu, ``Fairmot: On the fairness of detection and re-identification in multiple object tracking,'' \emph{International Journal of Computer Vision}, vol. 129, pp. 3069--3087, 2021.

\bibitem{deform_conv}
\BIBentryALTinterwordspacing
X.~Zhu, H.~Hu, S.~Lin, and J.~Dai, ``Deformable convnets v2: More deformable, better results,'' \emph{CoRR}, vol. abs/1811.11168, 2018. [Online]. Available: \url{http://arxiv.org/abs/1811.11168}
\BIBentrySTDinterwordspacing

\bibitem{Dota_dataset}
\BIBentryALTinterwordspacing
G.~Xia, X.~Bai, J.~Ding, Z.~Zhu, S.~J. Belongie, J.~Luo, M.~Datcu, M.~Pelillo, and L.~Zhang, ``{DOTA:} {A} large-scale dataset for object detection in aerial images,'' \emph{CoRR}, vol. abs/1711.10398, 2017. [Online]. Available: \url{http://arxiv.org/abs/1711.10398}
\BIBentrySTDinterwordspacing

\bibitem{8237586}
T.-Y. Lin, P.~Goyal, R.~Girshick, K.~He, and P.~Dollár, ``Focal loss for dense object detection,'' in \emph{2017 IEEE International Conference on Computer Vision (ICCV)}, 2017, pp. 2999--3007.

\end{thebibliography}
\end{document}